# Across the Levels of Analysis: Explaining Predictive Processing in Humans Requires More Than Machine-Estimated Probabilities

Sathvik Nair[1,2] & Colin Phillips[3,1]

[1]*University of Maryland Department of Linguistics,* [2]*University of Maryland Institute for Advanced Computer Studies,* [3]*Oxford University Faculty of Linguistics, Philology and Phonetics*

sathvik@umd.edu, colin.phillips@ling-phil.ox.ac.uk

**Abstract**

Under the lens of Marr's levels of analysis, we critique and extend the authors' two points about language models (LMs) and language processing: first, predicting upcoming linguistic information based on context is key to language processing, and second, that many advances in psycholinguistics would be impossible without LLMs. We also outline directions combining LLMs' strengths with psycholinguistic models.

In the 1960s, psycholinguists were excited about the possibility of a **computational-level** account of language processing where measures of "processing difficulty" were linked to aggregate measures of linguistic complexity. In that case, the difficulty measure (ratings or reading times for whole sentences) and the complexity measure (syntactic transformations) reflected the available (psycho)linguistic technologies of the time (Miller & McKean, 1964). The so-called Derivational Theory of Complexity (DTC) is widely cited as a cautionary tale about What Not to Do in cognitive science, but there are some enduring results establishing expectations for relative processing difficulty among different syntactic constructions.

Fifty to sixty years later, we again find enthusiasm for a computational-level account of language processing, one that links overall measures of processing difficulty to aggregate measures of linguistic complexity. Although today's technical advances operate at (and below) words instead of sentences, reflect much richer processing data (eye fixation durations, electrical potentials in the brain, or utterance latencies) and linguistic complexity (contextual probabilities estimated by LLMs), the core idea retains the half-century-old notion that a general-purpose measure to explain processing difficulty is a useful construct, now expressed as surprisal theory (Levy, 2008). With this in mind, it's important to learn from the mistakes of the DTC.

Just as in the 1960s, there are more cases where the general theory falls short, but there are some core insights that the field is reluctant to abandon. In the 2020s, these insights pertain to predictability effects in language processing (Staub, 2025); unlikely information in a given context is more difficult to process. Even simple LMs quantitatively describe these effects well, which has led researchers to claim language processing is associated with probabilistic inference (Smith & Levy,

2013). LLMs have since characterized predictability effects across different languages (Wilcox et al., Xu et al., 2023) and larger datasets (Shain et al., 2024), but are largely based on the same arguments. A major issue in this debate is that for LMs, predictability effects are *exclusively* linked to probabilistic inference, while the same cannot necessarily be said for humans, thanks to insights from the last few decades of psycholinguistic research (Brothers & Kuperberg, 2021; Szewczyk & Federmeier, 2022).

From the fall of the Derivational Theory of Complexity to the rise of surprisal theory, psycholinguists discovered finer-grained phenomena and experimental measures, largely shifting their attention to **algorithmic-level** questions. This fueled decades of research, and so few considered psycholinguistics to be equipped to make computational-level contributions. Futrell & Mahowald do say language processing is highly incremental, but psycholinguists' goal has been seen as characterizing the *underlying mental computations* from each step of incremental processing.

Contrary to the authors' claims that psycholinguistics has been hampered by weak statistical models of language, the field has made progress through empirical investigations of phenomena testing the processing system in tightly controlled experimental settings. Some examples include studies on linguistic illusions (Phillips et al., 2011; Paape, 2024), consequences of violated predictions (Frisson et al., 2017; Kuperberg et al., 2020), time delays (Tabor & Hutchins, 2004; Chow et al., 2018) and production-comprehension contrasts (Kandel et al., 2024; Lee & Phillips, 2025). Simpler, interpretable models, rather than LLMs, have proven useful for understanding these phenomena (Fitz & Chang, 2019; Aurnhammer et al., 2021; Nakamura et al., 2024).

That said, an important theoretical contribution of LLMs is to highlight the limits of predictability as a satisfying explanation for finer-grained aspects of language processing that unfold in time. LLM probabilities have limits as a functional explanation of human neural responses because they conflate semantic association with contextual expectations (Krieger et al., 2025). On the behavioral side, they insufficiently capture qualitative patterns in argument role reversals (Lee et al., 2024) and other types of linguistic illusions (Zhang et al., 2023), and quantitative patterns in syntactic processing (Huang et al., 2024).

These recent findings mark a shift from computational-level probabilities back towards algorithmic level processes. To this end, Futrell & Mahowald suggest imposing cognitively realistic resource constraints on LLMs' predictability estimates, and Giulianelli et al. (2023, 2024) derive LM-based difficulty measures for different levels of linguistic representation. We suggest another direction for progress: building on interactivity across levels of representation and incorporating predictability-based factors in process models. The field is moving in this direction already; LLMs complement insights about the time course of syntactic processing (Slaats et al., 2024; Timkey et al., 2025), can be combined with syntactic and semantic measures (Stanojević et al., 2023; De Santo, 2025; Jacobs et al., 2025), and support algorithmic-level accounts of inference (Vigly et al., 2025)

Process models can also be motivated by **implementational-level** ideas such as predictive coding (Rao & Ballard, 1999; Friston, 2009; Clark, 2013), grounded in theories of neurobiological function. Even if Futrell & Mahowald cite this work, LLMs' successes are at best rhetorically connected with predictive coding due to their limited biological plausibility (Schrimpf et al., 2021; Goldstein et al., 2022; Ryskin & Nieuwland, 2023).

Building on successfully implemented predictive coding models in other cognitive domains, their recent extension to linguistic data is promising. Not only do they elegantly explain disparate effects in controlled experimental settings (Nour Eddine et al., 2024), they also explain naturalistic reading data, with predictive power comparable to LMs with similar size (Ohams et al., 2026). Crucially, both these advances were derived from connecting simpler cognitive models (Rumelhart & McClelland, 1981; Elman, 1991) to biologically plausible methods of computation and provide interpretable measures that are directly tied to interactive cognitive processes, unlike LLMs.

To summarize, LMs have "revived" an older psycholinguistic approach, now providing far more accurate estimates of aggregate processing difficulty and linguistic complexity. Probabilistic models of linguistic data are useful, but progress in psycholinguistics will also require mechanistically characterizing the mental computations at the algorithmic level, in order to translate estimates of *which* words are (un)predictable from above into neural processes of *how* they are processed from below. Thus, psycholinguists can find different ways of loving what LMs can do for them, either as models of the core mechanism in language processing, or as components of more articulated algorithms, but their worries are justified.

## Acknowledgements

We are grateful to Samer Nour Eddine, Philip Resnik, Ziying Zhang, Matt Husband, and Alba Jorquera Jiménez de Aberásturi for valuable discussions. Sathvik Nair is supported by a NSF GRFP under Grant No. DGE 2236417.